\documentclass[10pt,twocolumn,letterpaper]{article}

\usepackage[pagenumbers]{cvpr} 
\usepackage{overpic}

\usepackage[accsupp]{axessibility}  
\usepackage[hyphens]{url}
\usepackage[pagebackref,breaklinks,colorlinks]{hyperref}
\usepackage[dvipsnames]{xcolor}  
\usepackage{colortbl}
\usepackage{graphicx}
\usepackage{amsmath}
\usepackage{amssymb}
\usepackage{booktabs}
\usepackage{multirow}
\usepackage{cuted}
\usepackage{bm}
\usepackage{xspace}
\usepackage{caption}
\usepackage{subcaption}
\usepackage{balance}
\usepackage{pifont}
\usepackage[normalem]{ulem}
\usepackage{arydshln}
\usepackage{lipsum}

\usepackage[capitalize]{cleveref}

\usepackage{xcolor}

\definecolor{applegreen}{rgb}{0.35, 0.81, 0.0}
\definecolor{atomictangerine}{rgb}{0.91, 0.45, 0.32}

\def\cvprPaperID{9161} 
\def\confName{CVPR}
\def\confYear{2023}

\begin{document}

\title{Blowing in the Wind: CycleNet for Human Cinemagraphs from Still Images}

\author{Hugo Bertiche$^{1,2}$ \and Niloy J. Mitra$^{3,4}$ \and Kuldeep Kulkarni$^4$ \and Chun-Hao Paul Huang$^4$ \and Tuanfeng Y. Wang$^4$ \and Meysam Madadi$^{1,2}$ \and Sergio Escalera$^{1,2}$ \and Duygu Ceylan$^4$\vspace{.25cm}
\and\normalsize $^1$Universitat de Barcelona 
\and\normalsize $^2$Computer Vision Center 
\and\normalsize $^3$University College London 
\and\normalsize $^4$Adobe Research
}


\twocolumn[{%
\renewcommand\twocolumn[1][]{#1}%
\maketitle
\thispagestyle{empty}
\begin{center}
    \includegraphics[width=\textwidth]{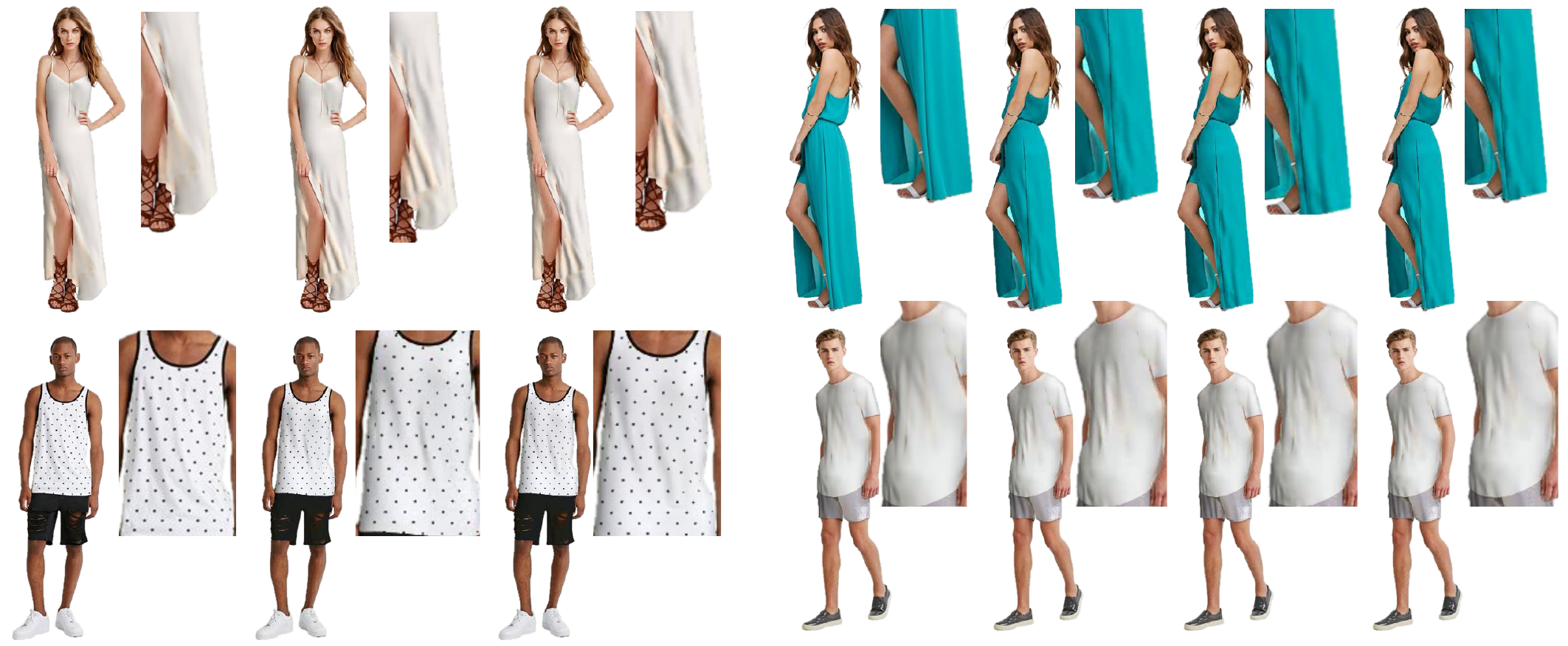}
    \captionof{figure}{We introduce a method for automatic human cinemegraph generation from single RGB images focusing on generating plausible garment animation as if they are blown in the wind. Please refer to the supplemental  videos for the animations.}
    \label{fig:teaser}
\end{center}
}]

\begin{abstract}
\vspace{-.25cm}
Cinemagraphs are short looping videos created by adding subtle motions to a static image. This kind of media is popular and engaging. However, automatic generation of cinemagraphs is an underexplored area and current solutions require tedious low-level manual authoring by artists. In this paper, we present an automatic method that allows generating human cinemagraphs from single RGB images. We investigate the problem in the context of dressed humans under the wind. At the core of our method is a novel cyclic neural network that produces looping cinemagraphs for the target loop duration. To circumvent the problem of collecting real data, we demonstrate that it is possible, by working in the image normal space, to learn garment motion dynamics on synthetic data and generalize to real data. We evaluate our method on both synthetic and real data and demonstrate that it is possible to create compelling and plausible cinemagraphs from single RGB images.
\end{abstract}

\section{Introduction}
\label{sec:intro}

Cinemagraph, a term originally coined by Jamie Beck and Kevin Burg, refers to adding dynamism to still images by adding \textit{minor and repeated movements}, forming a motion loop, to a still image. Such media format is both engaging and intriguing, as adding a simple and subtle motion can bring images to life. Creating such content, however, is challenging as it would require an artist to first set up and capture a suitable video, typically using a tripod, and then carefully mask out most of the movements in a post-processing stage.

We explore the problem of creating human cinemagraphs directly from a single RGB image of a person.  
Given a dataset of images and corresponding animated video pairs, a straightforward solution would be to train a fully supervised network to learn to map an input image to a plausible animated sequence. However, collecting such a dataset is extremely challenging and costly, as it would require capturing hundreds or thousands of videos of people holding a perfectly still pose under the influence of the wind from different known directions. While it is possible to simulate different wind force directions using oscillating fans in a lab setup~\cite{Bouman2013}, capturing the variability of garment geometry and appearance types in such a controlled setting is far from trivial. Hence, we explore the alternative approach of using synthetic data where different wind effects can easily be replicated using physically-based simulation. The challenge, then, is to close the synthetic-to-real gap, both in terms of garment dynamics and appearance variations.

We propose to address this generalization concern by operating in the gradient domain, i.e., using surface normal maps. Specifically, being robust to lighting or appearance variations, surface normals are arguably easier to generalize from synthetic to real, compared to RGB images. Moreover, surface normals are indicative of the underlying garment geometry (i.e., folds and wrinkles) and hence provide a suitable representation to synthesize geometric and resultant appearance variations~\cite{Lahner_2018_ECCV,zhang2021deep} as the garment interacts with the wind.

Further, we make the following technical contributions. First, we propose a novel \emph{cyclic} neural network formulation that directly outputs looped videos, with target time periods, without suffering from any temporal jumps. Second, we demonstrate how to condition the model architecture using wind parameters (e.g., direction) to enable control at test time. Finally, we propose a normal-based shading approach that takes the intermediate normals under the target wind attributes to produce RGB image frames. In Figure~\ref{fig:teaser}, we show that our method is applicable to a variety of real test images of different clothing types.

We evaluate our method on both synthetic and real images and discuss ablation results to evaluate the various design choices. We compare our approach against alternative approaches \cite{mahapatra2022controllable,wang2022latent} using various metrics as well as a user study to evaluate the plausibility of the generated methods. Our method achieves superior performance both in terms of quantitative metrics as well as the perceptual user study.

\section{Related Work}
\subsection{Looping video generation} In this work, we are interested in synthesizing \emph{cinemagraph} style looping animations where only certain parts of a frame are in motion. A typical method for creating such looping clips is to leverage video as input. Many approaches exist that solve an optimization problem to identify segments and transition points in the input video that can be looped seamlessly~\cite{schodl2000video, Bhat2004,Agarwala2005, couture2011panoramic, Tompkin2011,bai2012selectively, Yeh2012, Liao2013, Liao2015}. While we focus on generating such a looping clip from a static single image, we use a video based method~\cite{Liao2015} to ensure our training data is looped properly.

In the context of animating a single image in a looping manner, one approach is to warp regions of the image using Fourier methods in a stochastic manner which amounts to displacing the original texture~\cite{Chuang2005}. Another approach is to transfer the phase patterns from an example video to the given input image~\cite{Prashnani2017}. Okabe et al.~\cite{Okabe2009} also transfer the motion patterns from an example video to an input image of a fluid. Specifically, they map the example video to a constant flow and residual layers, which represent the high frequency motion patterns that are not explained by warping a reference frame using constant flow. Such residual patterns are transferred to the input image. These methods work best for natural phenomena such as water and fire where flow-based texture displacement and warping result in plausible animation. Halperin et al.~\cite{halperin2021endless} present another approach to animating a single image by focusing on repeating patterns. While demonstrating impressive results, such a method is not suitable for our problem since the motion a garment undergoes blowing in the wind is fundamentally different than displacing repeating patterns.

With the recent success of deep learning methods, several learning based approaches have been proposed to create looping animations from single images. While Endo et al.~\cite{Endo2019} predict a flow map to warp the input images directly, Holynski et al.~\cite{Holynski_2021_CVPR} first generate a constant flow map directly from a single image and then warp image features using the generated flow map to synthesize the RGB frames. In a follow-up work, Mahapatra et al.~\cite{mahapatra2022controllable} extend this framework to provide additional control of the motion direction and region of the image to be animated. We compare our method to this state-of-the-art approach and show that the assumption of constant flow is not suitable for garment motion and leads to unsatisfactory results. Recently, Fan et al.~\cite{fan2022simulating} present a method to animate fluids in a still image. Their method uses an additional depth map estimation to generate a surface mesh for the fluid region and thus utilizes physically based simulation priors to predict a motion field. While our approach of incorporating a surface normal map representation is similar, we focus on very different types of motions in our work.

\begin{figure*}
    \centering
    \includegraphics[width=\textwidth]{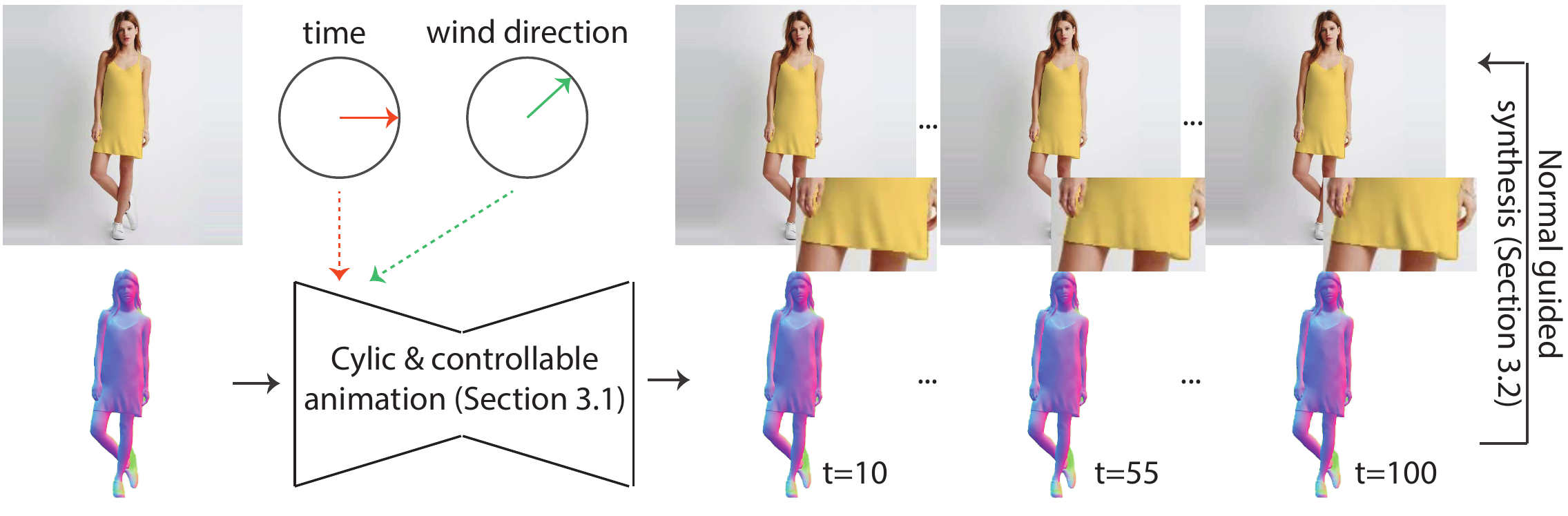}
    \caption{Given an input image (top, left) and its predicted surface normal map (bottom, left), we present a network that synthesizes a set of surface normals that resemble the effect of the garment blowing in the wind with a given direction. We ensure a looped animation by encoding the time $t$ with a cyclic positional encoding with respect to a predefined loop duration (150 frames in our experiments). We then synthesize the corresponding RGB images demonstrating plausible garment deformation using an intrinsic image decomposition technique.}
    \label{fig:pipeline}
\end{figure*}

\subsection{Animating single images} With the success of deep learning, several methods have been recently proposed  to animate a given image. One approach is based on using a driving video and focus on synthesizing specific type of content and motion such as time-lapse videos~\cite{cheng2020time,Logacheva_2020_ECCV}, facial and body animation~\cite{Siarohin_2019_NeurIPS,wang2022latent}. We compare our method to the most recent method of Wang et al.~\cite{wang2022latent} and show that it is not suitable to capture the subtle motions observed in a human cinemagraph.

Another line of work directly predicts video or future frames from a given single image~\cite{visualdynamics16,Li2018,Xiong_2018_CVPR,dtvnet} or a semantic map~\cite{pan2019video}. Dorkenwald et al.~\cite{dorkenwald2021stochastic} learn a generative model that encodes a latent residual representation and sample such 
latent code to synthesize a video from a given image. Many of these methods, however, synthesize multiple frames at the same time and hence operate only at low resolution without providing control. To address the latter challenge, Blatmann et al.~\cite{blattmann2021understanding, blattmann2021ipoke} enable the user to provide a poke that determines the final location of a sparse point in the input image. The resulting videos, however, are not looped in contrast to the type of animation we are interested in synthesizing.

Another interesting direction is to train a single image based generator~\cite{rottshaham2019singan}, which can then be utilized to generate animations by providing random walks of the appearance of the object of interest in the latent space. Arora et al.~\cite{arora2021singan} extend this approach to work with an input GIF. While impressive, such approaches do not provide the controllability we aim to achieve with our approach, however.

\section{Methodology}

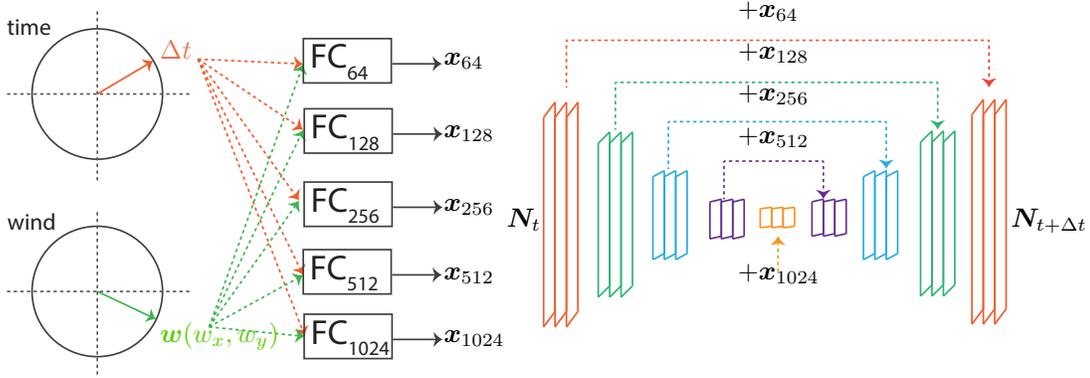
\begin{figure*}
\centering
    \begin{overpic}[width=.8\textwidth]{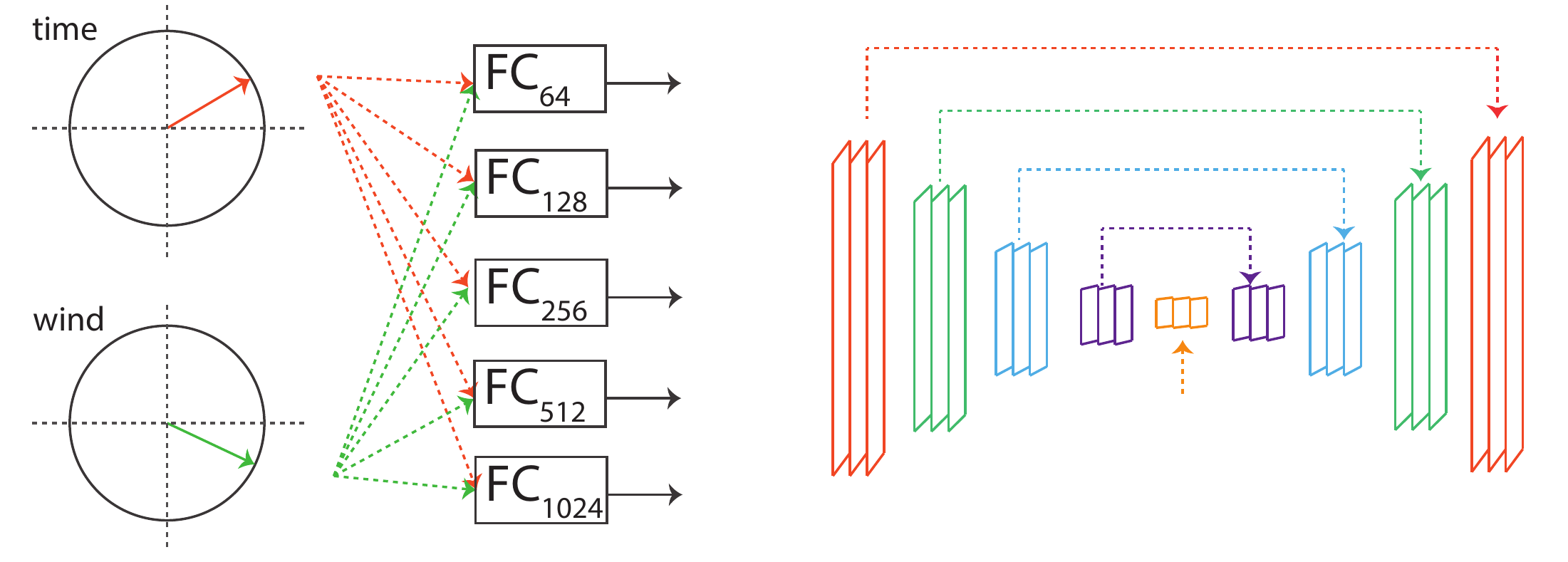}
    \put(16,31){
      \color{atomictangerine} $\Delta t$
      }
      \put(16,4){
      \color{applegreen} $\bm{w}(w_x,w_y)$
      }
        \put(43,30.5){
      \color{black} $\bm{x}_{64}$
      }
       \put(71,35){
      \color{black} $+\bm{x}_{64}$
      }
      \put(43,23.5){
      \color{black} $\bm{x}_{128}$
      }
      \put(71,31){
      \color{black} $+\bm{x}_{128}$
      }
      \put(43,16.5){
      \color{black} $\bm{x}_{256}$
      }
      \put(71,27){
      \color{black} $+\bm{x}_{256}$
      }
      \put(43,10){
      \color{black} $\bm{x}_{512}$
      }
       \put(71,23){
      \color{black} $+\bm{x}_{512}$
      }
      \put(43,4){
      \color{black} $\bm{x}_{1024}$
      }
      \put(71,10){
      \color{black} $+\bm{x}_{1024}$
      }

      \put(49,15){
      \color{black} $\bm{N}_t$
      }
      \put(97,15){
      \color{black} $\bm{N}_{t+\Delta t}$
      }
      
    \end{overpic}
    \caption{Cyclic wind-conditioned UNet. Given an input normal map $N_t$, a delta time increment $\Delta t$, and a wind direction $\bm{w}$; we extend the standard UNet architecture to give it a cyclic behaviour. We encode the time using a cylic positional encoding and concatenate with the wind direction. We pass the concatenated features through different fully convolutional layers to extract features of varying dimensions. The resulting features are provided as skip connections to the UNet architecture which synthesizes the final normal map $N_{t+\Delta t}$.}
    \label{fig:architecture}
\end{figure*}

Given a single input RGB image of a person, $\bm{I}\in\mathbb{R}^{W\times H\times3}$, our goal is to generate a looped video sequence, $\bm{V} := \{\bm{I}_0, \bm{I}_1, ..., \bm{I}_t | \bm{I}_0 = \bm{I}_t\}$, where the loose garments worn by the person exhibit a plausible motion as if blown in the wind. We assume the direction of the wind can be provided by a unit vector $\bm{w}$ in the image plane to control the output animation. Hence, our goal is to learn the mapping $\mathcal{F}(\bm{I},\bm{w}) \longrightarrow \bm{V}^{\bm{w}}$. 

To more effectively represent the underlying garment geometry and the changes it undergoes due to the wind force, our method operates on the surface normal map $\bm{N}$ that corresponds to the input image $\bm{I}$. Specifically, given an input image $\bm{I}$, we first predict the surface normal map using an off-the-shelf normal estimator~\cite{Bae2021}. We then propose a novel cyclic network architecture that maps $\bm{N}$ to a sequence of normal maps $\bm{V_N}^{\bm{w}} := \{\bm{N}_0, \bm{N}_1, ..., \bm{N}_t | \bm{N}_0 = \bm{N}_t\}$ that demonstrate plausible motion of the underlying garment under the influence of a wind force with a direction given by $\bm{w}$. Finally, we synthesize back the corresponding RGB images given the original input image and the sequence of animated normal maps using a constrained reshading approach. We provide the overall pipeline in Figure~\ref{fig:pipeline} and next discuss the details of our approach. 

%

\subsection{Cyclic and Controllable Animation}

Given an input normal map $\bm{N}$ and a wind direction $\bm{w}$, our goal is to learn the mapping $\mathcal{F_N}(\bm{N},\bm{w}) \longrightarrow \bm{V_N}^{\bm{w}} = \{\bm{N}_0, \bm{N}_1, ..., \bm{N}_t | \bm{N}_0 = \bm{N}_t\}$ where $\bm{V_N}^{\bm{w}}$ demonstrates plausible garment animation. Our goal is to synthesize a cyclic animation sequence with a predefined period of $T$, so that $t=T-1$. This amounts to synthesizing normal maps that satisfy constraints $\bm{N}_t = \bm{N}_{t+kT}$ $\forall k\in\mathbb{Z}$.

We tackle this problem as an image-to-image translation task where our goal is to learn $f(\bm{N}_t, \Delta t, \bm{w}) \longrightarrow \bm{N}_{t+\Delta t}$ where $\Delta t \in [-T/2, T/2]$. Note that, since we are interested in looped animations, negative values for $\Delta t$ correspond to valid animation samples. We realize the function $f$ as a UNet architecture that is conditioned on both the residual time $\Delta t$ and the wind direction $\bm{w}$, as shown in Figure~\ref{fig:architecture}. To enforce a cyclic behaviour, we first encode $\Delta t$ using sinusoidal functions as:
\begin{equation}
    \label{eq:time_encoding}
    \begin{split}
        \varphi_{\Delta t} &= \frac{2\pi n}{T}\Delta t,  \quad n = 1, 2, 3, 4, 5.\\
        \bm{x}_{\Delta t} &= \{\text{cos}(\varphi_{\Delta t}), \text{sin}(\varphi_{\Delta t})\}.      
    \end{split}
\end{equation}
This formulation ensures that $f(\bm{N}_t, \Delta t + kT, \bm{w})$ with $k\in\mathbb{Z}$ gives the same output resulting in a looping animation sequence. Similar to common practice in positional encoding~\cite{vaswani2017attention}, we observe that using multiples of the data frequency ($\omega = 2\pi n/T$) helps to learn higher frequency motions while still enforcing a global cyclic behaviour with period $T$. Note then how the time encoding $\bm{x}_{\Delta t}$ consists of multiple circumferences parameterized by $\Delta t$. 

We represent the wind direction as a unit vector as $\bm{w}$ in the image plane. We concatenate $\bm{w}$ with $\bm{x}_{\Delta t}$ resulting in the final conditioning code $\bm{x} := \bm{x}_{\Delta t} \| \bm{w} = (x_{\Delta t,0}, x_{\Delta t,1}, ..., x_{\Delta t,2n}, w_x, w_y)$. We condition the UNet by introducing $\bm{x}$ at each feature map extracted by the encoder at different scales. To do so, we first linearly transform $\bm{x}$ to the corresponding feature map dimensionality with learnable weights $\{\bm{W}_i\in\mathbb{R}^{F_i\times D}\}$, where $F_i$ is the number of channels of the $i$-th feature map and $D$ is the dimensionality of $\bm{x}$. We apply $1\times 1$ conovolutions to the feature maps before and after combining them with $\bm{x}$.

\subsection{Normal Guided Synthesis}
\label{sec:reshading}

The final stage of our approach focuses on computing the final cinemagraph $\bm{V}$ given the original input RGB image $\bm{I}$ and the predicted normal map sequence $\bm{V_N}^{\bm{w}}$. To this end, we rely on the concept of intrinsic image decomposition where we assume images can be decomposed into two layers $\bm{I} = \bm{S}\bm{R}$: (i) the reflectance $\bm{R}\in\mathbb{R}^{W\times H\times3}$, which denotes the albedo invariant color of the materials, and (ii) the shading $\bm{S}\in\mathbb{R}^{W\times H}$ which is the result of the interaction of the light with the underlying geometry of the garment. In particular, the shading layer is crucial in how we perceive the changes in the fold and wrinkle patterns of the garment as it is animated. Given this observation, we synthesize a new shading layer that is consistent with the animated surface normal maps. Then, when composed with the original reflectance map reflects it generates the intended animation.
%
%

Given the input image $\bm{I}$, we first run an off-the-shelf intrinsic image decomposition method~\cite{bell14intrinsic} to obtain the reflectance map $\bm{R}$ and the shading map $\bm{S}$. Assuming a simple lighting model composed of a directional and ambient light, we optimize for the light parameters using the predicted surface normal map from the input image:
\begin{equation}
    \bm{S} = \text{max}(0, -\bm{N}\bm{l}) + \delta,
    \label{eq:shading}
\end{equation}

where $\bm{l}\in\mathbb{R}^3$ is the light direction and $\delta\in\mathbb{R}^+$ is the ambient light. Given the predicted animated surface normal map sequence $\bm{\hat{V}_N}$, we generate a new shading map sequence and composite it with the original reflectance map $\bm{R}$ to obtain the final RGB sequence $\bm{\hat{V}}$. At inference time, the user is required to provide a mask to denote the region of interest where motion is desired to be synthesized. Hence, we composite the original image and the synthesized RGB images based on this mask to provide the final output. While this approach changes only the shading without actually warping the texture of the garment, it is sufficient to provide the perception of a plausible animation.

\vspace{-.625cm}
\textcolor{black}{
\paragraph{Local vs Global.} We design this methodology so it leans towards a local solution. The reasons for this are as follows. On one hand, cinemegraphs are characterized by subtle motions (local). On the other hand, \textit{local} solutions generalize better, which is specially important for our approach to handle real test samples from a synthetic training set.
}

\section{Experiments}

In the following section, we describe the experimental setup and the qualitative and quantitative results. We detail the data used for training and evaluation, define the metrics, and briefly introduce the state-of-the-art baselines used for comparison. Finally, we provide a discussion of the results.

\begin{figure}
    \centering
    \includegraphics[width=\columnwidth]{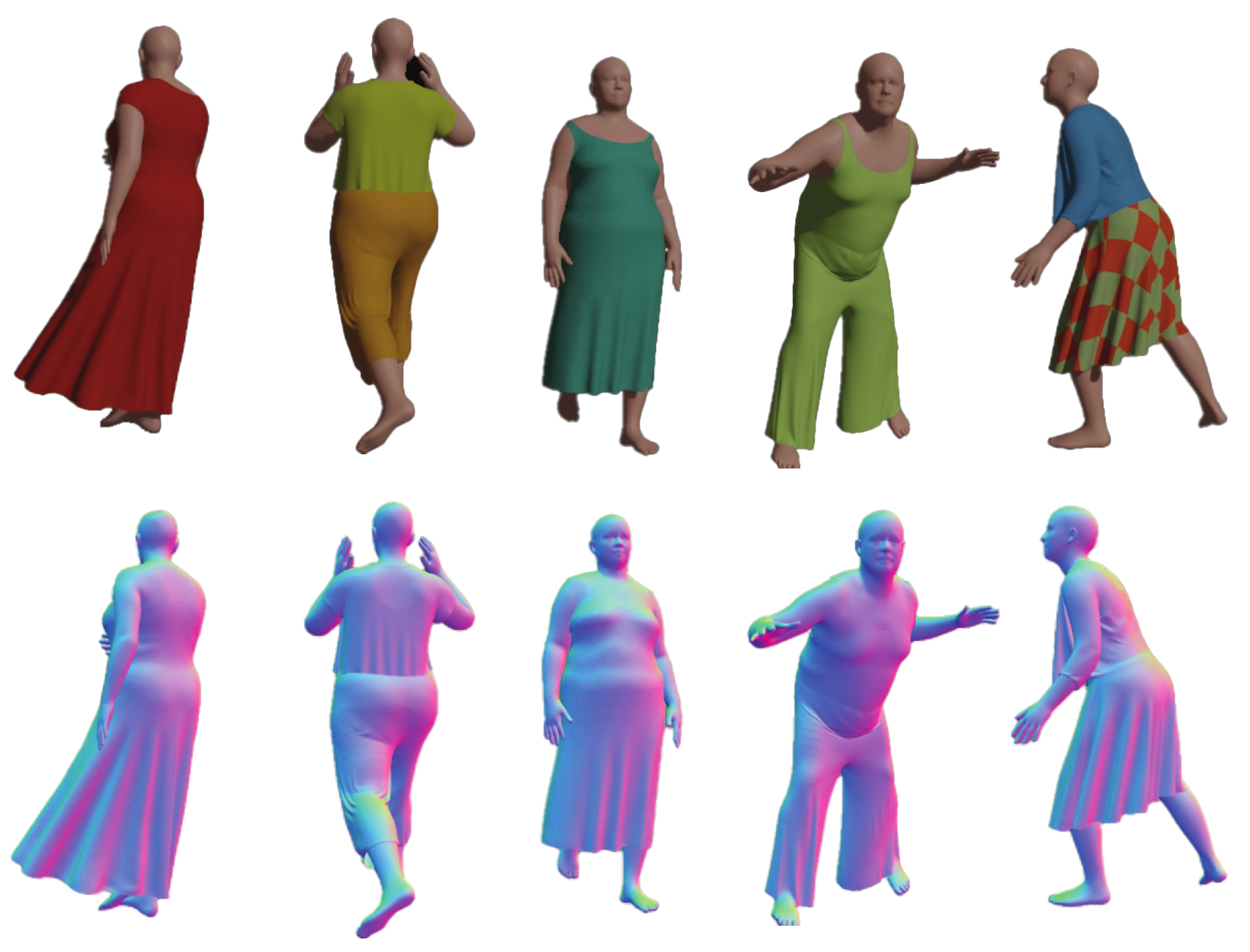}
    \caption{We train ours on a synthetic dataset that consists of different garment types draped on bodies with varying shape and poses acquired from the Cloth3D dataset~\cite{Bertiche2020}. We simulate the effect of wind and render  the corresponding RGB and surface normal images.}
    \label{fig:synthetic}
\end{figure}

\begin{figure}
    \centering
    \begin{subfigure}{.24\linewidth}
        \centering
        \includegraphics[width=\textwidth]{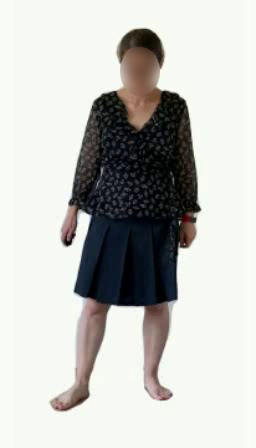}
    \end{subfigure}
    \begin{subfigure}{.24\linewidth}
        \centering
        \includegraphics[width=\textwidth]{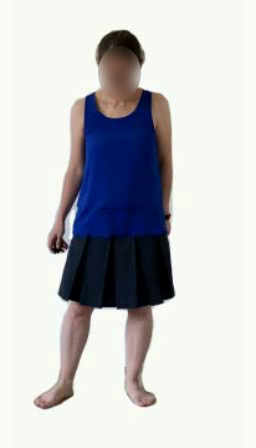}
    \end{subfigure}
    \begin{subfigure}{.24\linewidth}
        \centering
        \includegraphics[width=\textwidth]{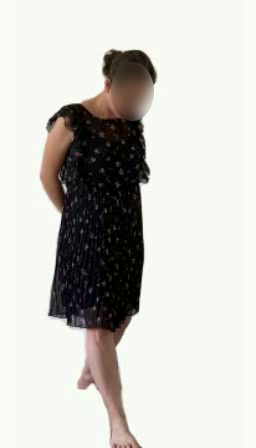}
    \end{subfigure}
    \begin{subfigure}{.24\linewidth}
        \centering
        \includegraphics[width=\textwidth]{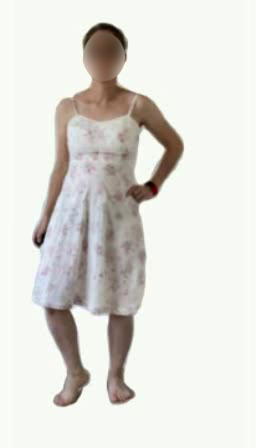}
    \end{subfigure}
    \begin{subfigure}{.24\linewidth}
        \centering
        \includegraphics[width=\textwidth]{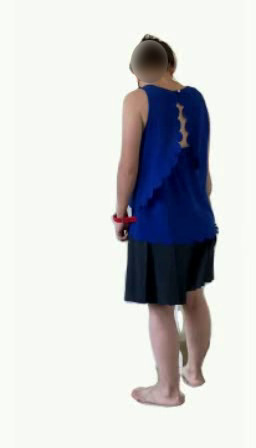}
    \end{subfigure}
    \begin{subfigure}{.24\linewidth}
        \centering
        \includegraphics[width=\textwidth]{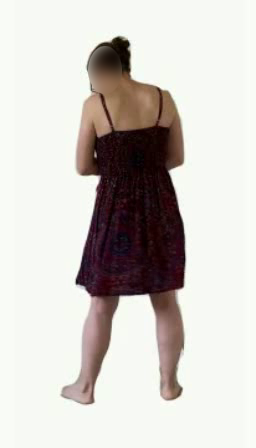}
    \end{subfigure}
    \begin{subfigure}{.24\linewidth}
        \centering
        \includegraphics[width=\textwidth]{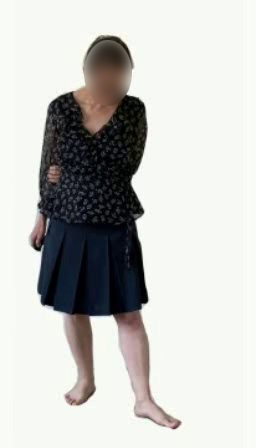}
    \end{subfigure}
    \begin{subfigure}{.24\linewidth}
        \centering
        \includegraphics[width=\textwidth]{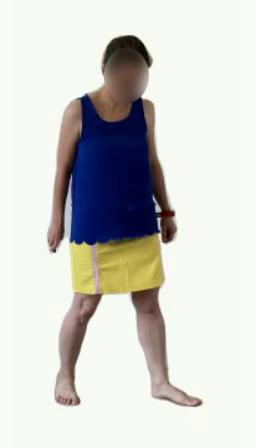}
    \end{subfigure}

    \caption{We capture a small real dataset where the subject keeps a still pose during the sequence while a fan generates wind. Different garment types show different dynamics.}
    \label{fig:real_capture} 
\end{figure}

\paragraph{Datasets.}
In order to train our network, we generate a synthetic dataset that consists of different type of garments draped on human bodies with varying shape and pose. Specifically, we sample human body and garment pairs from the Cloth3D dataset~\cite{Bertiche2020}, which is a large-scale dataset of clothed 3D humans. We select $1500$ samples with skirt and dresses and $500$ samples with other clothing types (e.g., trousers, tshirts). Each sample in the original Cloth 3D dataset is a motion sequence. We randomly choose one of the frames in each sequence as a random human body pose. The chosen frame, body and outfit, defines the initial conditions of our cloth simulation. We use Blender\cite{blender} to run the simulations. To this end, we choose a random wind direction in the image plane with constant wind force, and simulate the cloth dynamics while the underlying body remains still. Each simulation output is rendered from a fixed viewpoint with a predefined lighting setup. We apply random checkerboard texture patterns to some garments and assign a uniform color material to others. In addition to RGB output, we also render the corresponding surface normal maps and segmentation masks (body, cloth and background). Figure~\ref{fig:synthetic} shows examples from our dataset. We simulate each sample for $250$ frames at $30$ fps. We observe that the garment drapes on the body in roughly the first $50$ frames of the sequence and later starts blowing in the wind. It is not trivial to guarantee the resulting garment animation is cyclic in such a physically based simulation setup. Hence, we process the resulting animations with the method of Liao et al.~\cite{Liao2015} which detects loops in an input video. After this step, we obtain animation sequences of length $150$ frames which we use as the duration of loops, i.e., $T=150$.

In addition to synthetic data, we test our method on real samples from the Deep Fashion dataset~\cite{liuLQWTcvpr16DeepFashion} as well as additional stock images to test generalization. To evaluate if the predictions obtained on real samples contain plausible cloth dynamics, we capture a small set of real examples. Specifically, we ask a human subject wearing different types of garments to hold a still pose next to an oscillating fan while we record a short video sequence with a fixed camera mounted on a tripod. We record 50 such videos demonstrating 8 different outfit types. Similar to synthetic data, we process each video with the method of Liao et al.~\cite{Liao2015} to obtain looped animations. Figure~\ref{fig:real_capture} shows some real samples.

\paragraph{Evaluation Metrics.}
We evaluate our method and baselines on synthetic data where we can access ground truth image and animation pairs. First, we adopt metrics that focus on pixel-level similarity. Specifically, we report per-pixel mean average error~(MAE), mean squared error~(MSE), root of mean squared error~(RMSE), and PSNR. In addition we report metrics that focus on more structural (SSIM~\cite{wang2004image}) and perceptual similarities (LPIPS~\cite{zhang2018perceptual}). For DeepFashion samples we do not have ground truth video data. Hence, in order to evaluate the plausibility of the generated animated sequences we use Frechet Video Distance~(FVD \cite{unterthiner2018towards}) against the real data we have captured.


\paragraph{Baselines.}
We compare our method to two baselines. First, we compare with the work of Mahapatra et al.~\cite{mahapatra2022controllable}, which extends the original Eulerian motion fields approach~\cite{holynski2021animating} to a controllable setup. Since this method is a flow based approach and uses optical flow information to be provided in the dataset, we train it with the looped RGB videos in our synthetic dataset where optical flow can be more reliably estimated using off-the-shelf methods~\cite{teed2020raft}. For each looped sequence, we extract a mask denoting the region where motion is observed and a sparse set of motion directions from the estimated optical flow. We also compare our method to LIA~\cite{wang2022latent}, a state-of-the-art single image-based controllable video generation framework. Since LIA requires a target video sequence to specify the desired animation, we provide the ground truth animation sequences as targets both during training and testing. While it is not possible to use this configuration in a real setup, it provides the best possible results. Outperforming LIA under this configuration means outperforming it under any other.

\begin{figure}[!ht]
    \centering
    \rotatebox{90}{\quad LIA\cite{wang2022latent}}
    \begin{subfigure}{.22\linewidth}
        \centering
        \includegraphics[width=\linewidth]{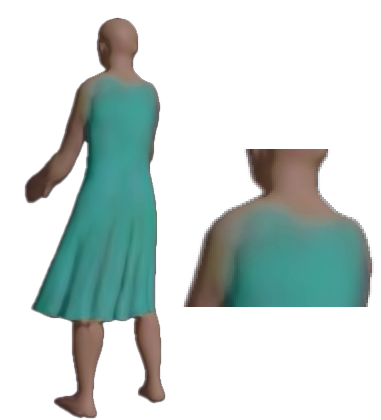}
    \end{subfigure}
    \begin{subfigure}{.22\linewidth}
        \centering
        \includegraphics[width=\linewidth]{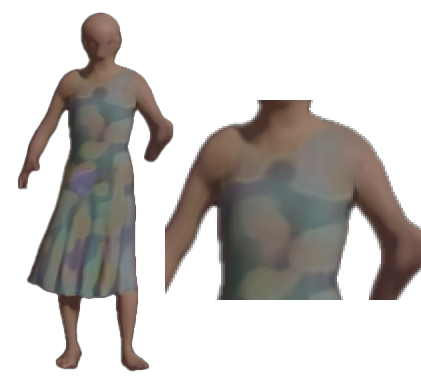}
    \end{subfigure}
    \begin{subfigure}{.22\linewidth}
        \centering
        \includegraphics[width=\linewidth]{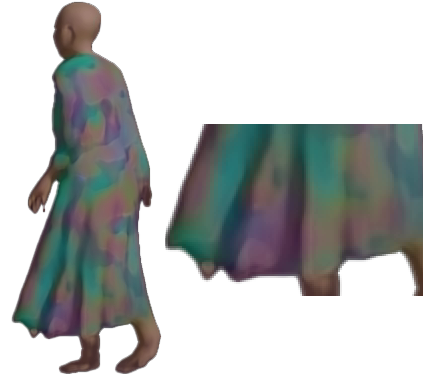}
    \end{subfigure}
    \begin{subfigure}{.22\linewidth}
        \centering
        \includegraphics[width=\linewidth]{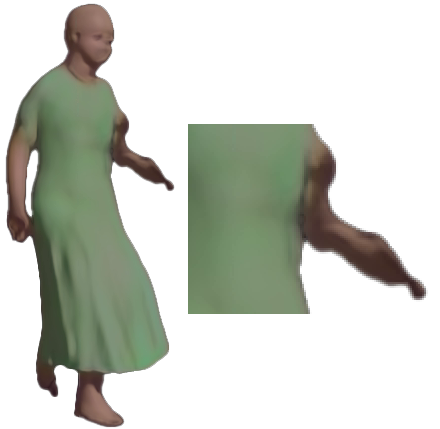}
    \end{subfigure}
    \\
    \hspace{-.5cm}
    \rotatebox{90}{Controllable}
    \rotatebox{90}{\hspace{.075cm}Anim.\cite{mahapatra2022controllable}}
    \begin{subfigure}{.22\linewidth}
        \centering
        \includegraphics[width=\linewidth]{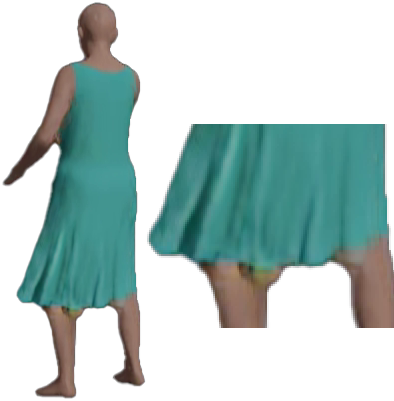}
    \end{subfigure}
    \begin{subfigure}{.22\linewidth}
        \centering
        \includegraphics[width=\linewidth]{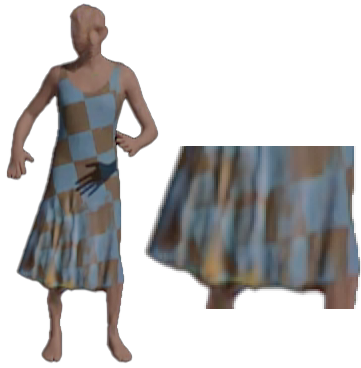}
    \end{subfigure}
    \begin{subfigure}{.22\linewidth}
        \centering
        \includegraphics[width=\linewidth]{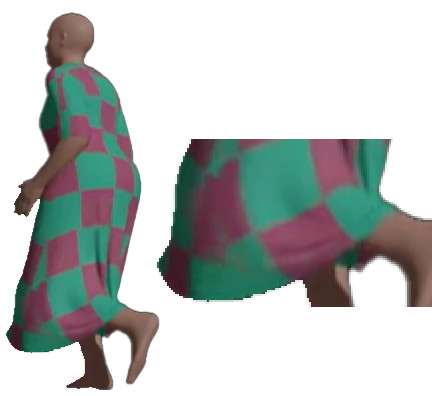}
    \end{subfigure}
    \begin{subfigure}{.22\linewidth}
        \centering
        \includegraphics[width=\linewidth]{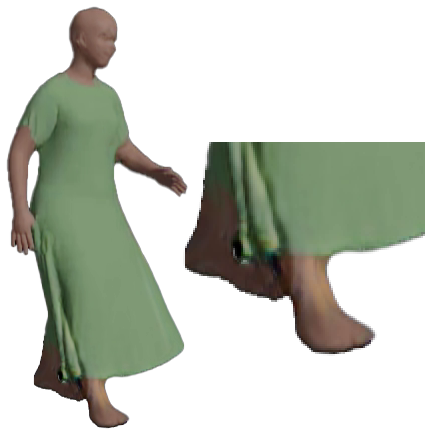}
    \end{subfigure}
    \\
    \rotatebox{90}{\quad\quad RGB}
    \begin{subfigure}{.22\linewidth}
        \centering
        \includegraphics[width=\linewidth]{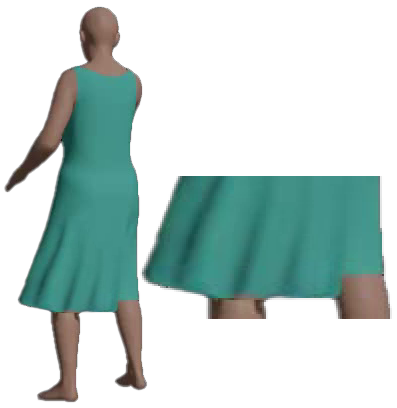}
    \end{subfigure}
    \begin{subfigure}{.22\linewidth}
        \centering
        \includegraphics[width=\linewidth]{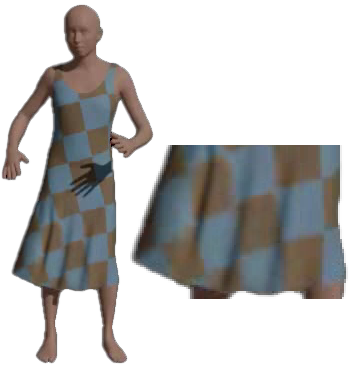}
    \end{subfigure}
    \begin{subfigure}{.22\linewidth}
        \centering
        \includegraphics[width=\linewidth]{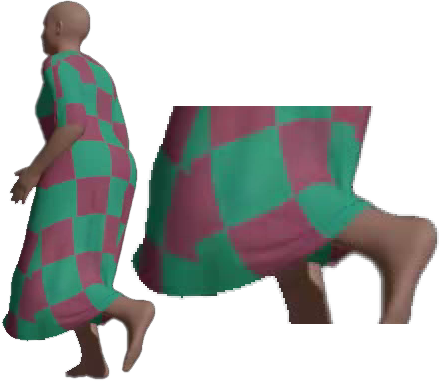}
    \end{subfigure}
    \begin{subfigure}{.22\linewidth}
        \centering
        \includegraphics[width=\linewidth]{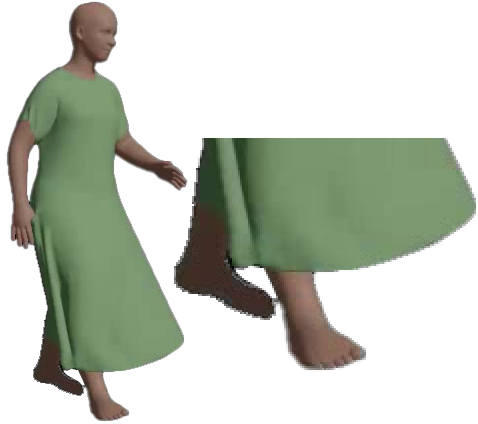}
    \end{subfigure}
    \\
    \rotatebox{90}{\quad Normals}
    \begin{subfigure}{.22\linewidth}
        \centering
        \includegraphics[width=\linewidth]{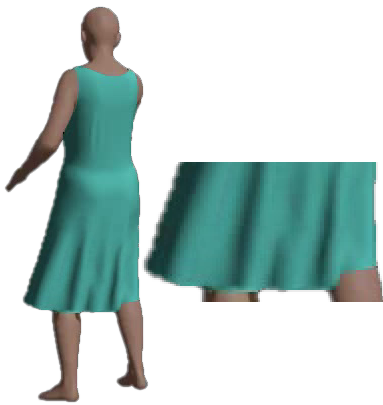}
    \end{subfigure}
    \begin{subfigure}{.22\linewidth}
        \centering
        \includegraphics[width=\linewidth]{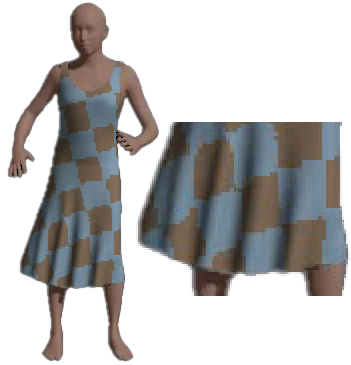}
    \end{subfigure}
    \begin{subfigure}{.22\linewidth}
        \centering
        \includegraphics[width=\linewidth]{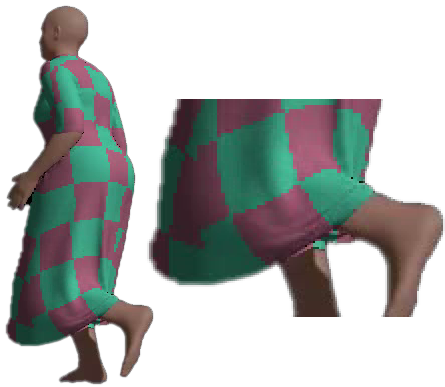}
    \end{subfigure}
    \begin{subfigure}{.22\linewidth}
        \centering
        \includegraphics[width=\linewidth]{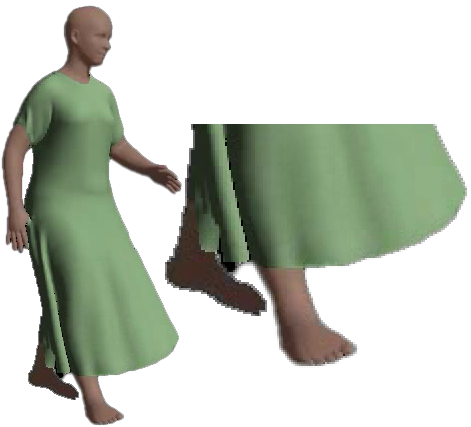}
    \end{subfigure}
    \\
    \hspace{-.5cm}
    \rotatebox{90}{Estimated}
    \rotatebox{90}{\hspace{.0375cm} Normals}
    \begin{subfigure}{.22\linewidth}
        \centering
        \includegraphics[width=\linewidth]{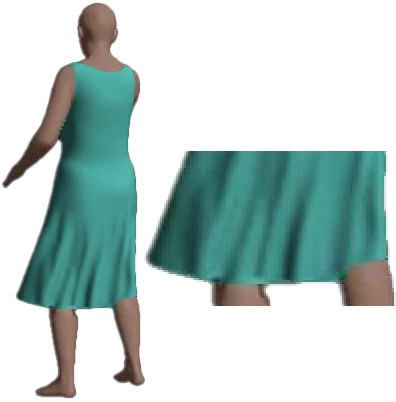}
    \end{subfigure}
    \begin{subfigure}{.22\linewidth}
        \centering
        \includegraphics[width=\linewidth]{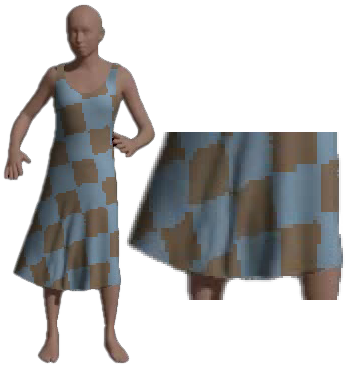}
    \end{subfigure}
    \begin{subfigure}{.22\linewidth}
        \centering
        \includegraphics[width=\linewidth]{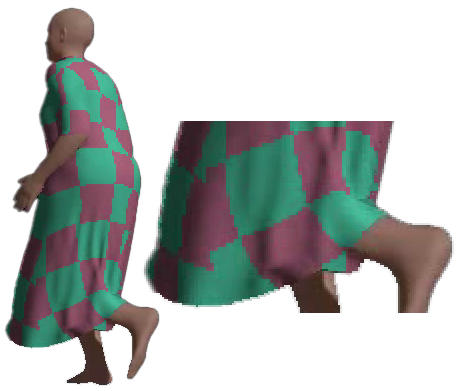}
    \end{subfigure}
    \begin{subfigure}{.22\linewidth}
        \centering
        \includegraphics[width=\linewidth]{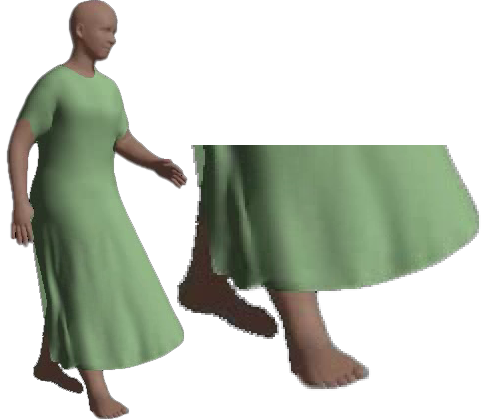}
    \end{subfigure}
    \\
    \rotatebox{90}{\quad\quad GT}
    \begin{subfigure}{.22\linewidth}
        \centering
        \includegraphics[width=\linewidth]{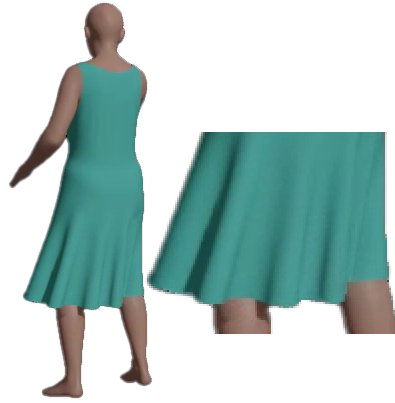}
    \end{subfigure}
    \begin{subfigure}{.22\linewidth}
        \centering
        \includegraphics[width=\linewidth]{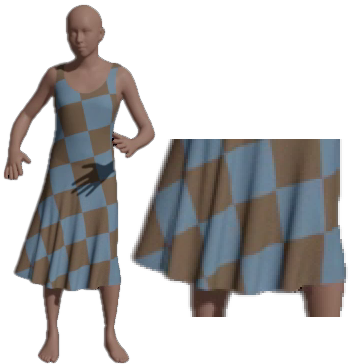}
    \end{subfigure}
    \begin{subfigure}{.22\linewidth}
        \centering
        \includegraphics[width=\linewidth]{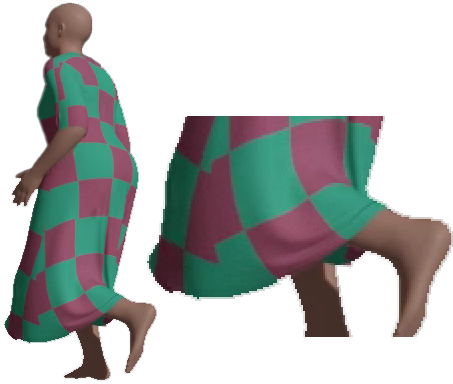}
    \end{subfigure}
    \begin{subfigure}{.22\linewidth}
        \centering
        \includegraphics[width=\linewidth]{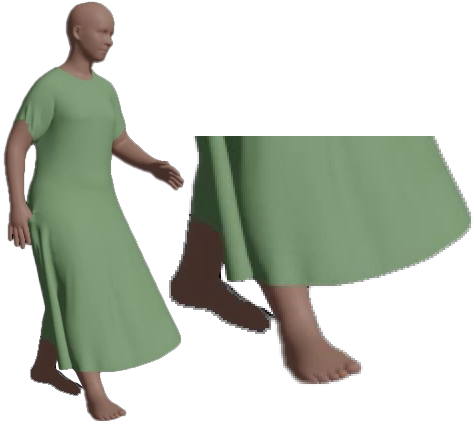}
    \end{subfigure}
    \caption{Ablation study. We evaluate qualitatively the image synthesis capacity of each methodology in our synthetic dataset, namely the state-of-the-art baselines LIA\cite{wang2022latent} and the work on controllable animations of \cite{mahapatra2022controllable} and the different variations of our approach, trained on different data modalities. Normal-based solutions require a re-shading step (Sec.~\ref{sec:reshading}). We observe SOTA solutions (warping-based) are sub-optimal for our setting. Ours is able to generate consistent images with plausible wrinkles.}
    \label{fig:ablation}
\end{figure}

\begin{figure*}[t!]
    \centering
    \begin{subfigure}{.19\linewidth}
        \centering
        \includegraphics[width=\textwidth]{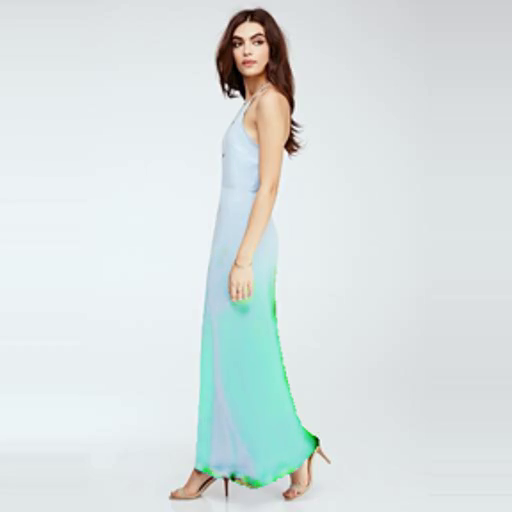}
    \end{subfigure}
    \begin{subfigure}{.19\linewidth}
        \centering
        \includegraphics[width=\textwidth]{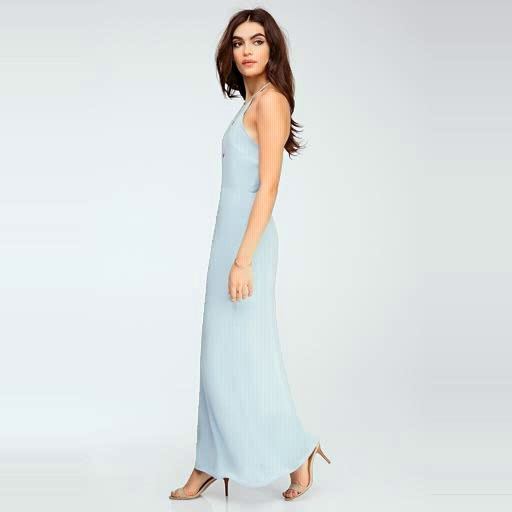}
    \end{subfigure}
    \begin{subfigure}{.19\linewidth}
        \centering
        \includegraphics[width=\textwidth]{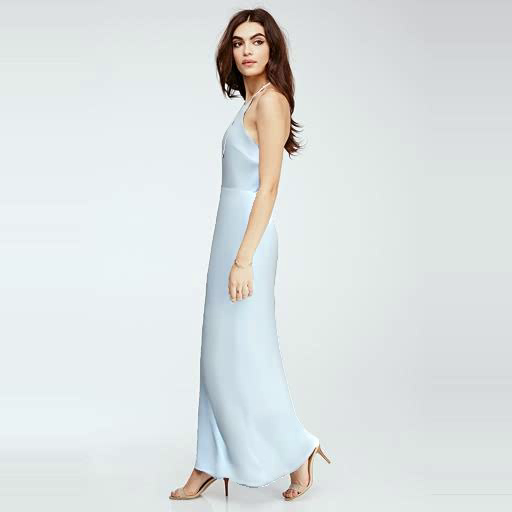}
    \end{subfigure}
    \begin{subfigure}{.19\linewidth}
        \centering
        \includegraphics[width=\textwidth]{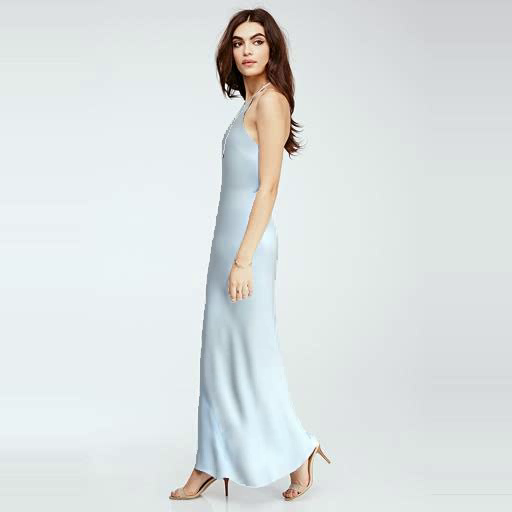}
    \end{subfigure}
    \begin{subfigure}{.19\linewidth}
        \centering
        \includegraphics[width=\textwidth]{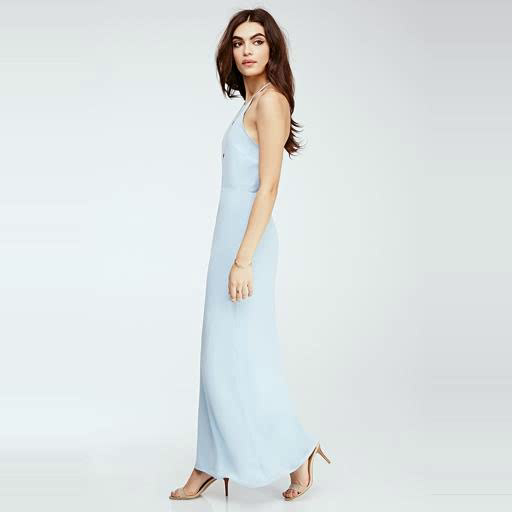}
    \end{subfigure}
    \begin{subfigure}{.19\linewidth}
        \centering
        \includegraphics[width=\textwidth]{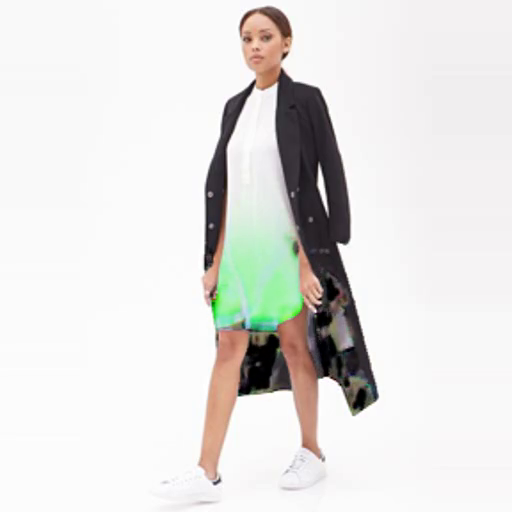}
    \end{subfigure}
    \begin{subfigure}{.19\linewidth}
        \centering
        \includegraphics[width=\textwidth]{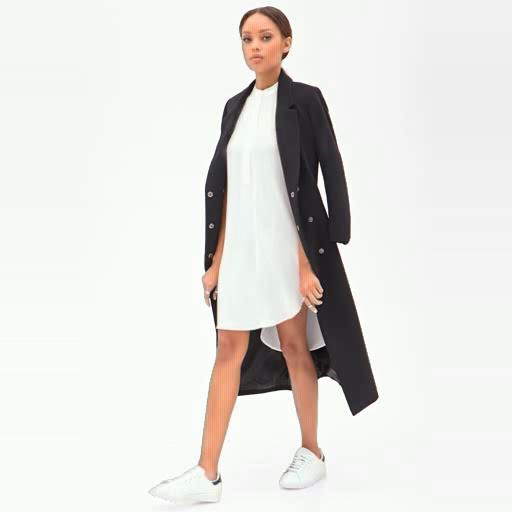}
    \end{subfigure}
    \begin{subfigure}{.19\linewidth}
        \centering
        \includegraphics[width=\textwidth]{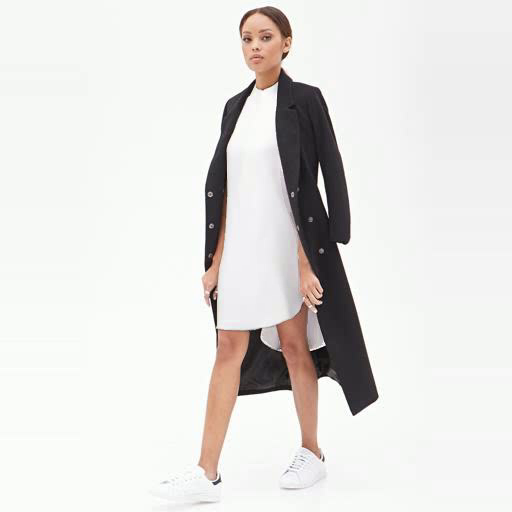}
    \end{subfigure}
    \begin{subfigure}{.19\linewidth}
        \centering
        \includegraphics[width=\textwidth]{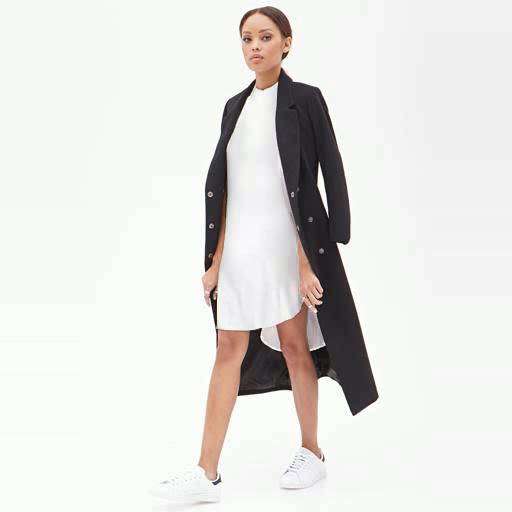}
    \end{subfigure}
    \begin{subfigure}{.19\linewidth}
        \centering
        \includegraphics[width=\textwidth]{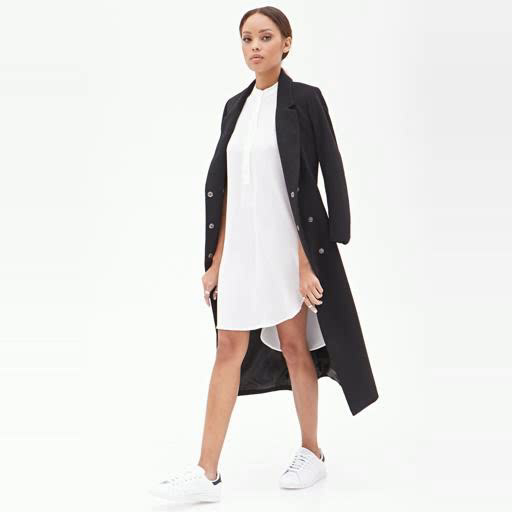}
    \end{subfigure}
    \begin{subfigure}{.19\linewidth}
        \centering
        \includegraphics[width=\textwidth]{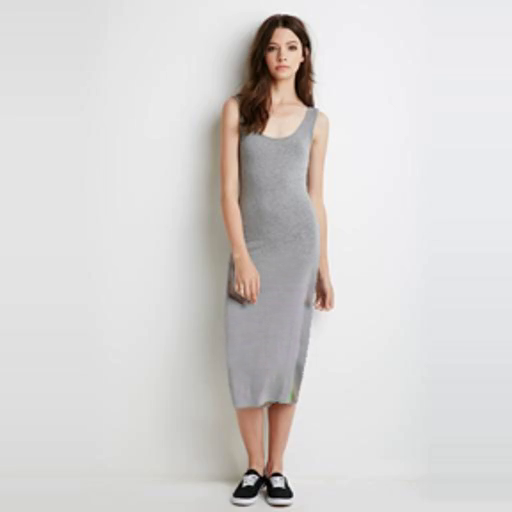}
        \caption{Contr. Anim.\cite{mahapatra2022controllable}}
    \end{subfigure}
    \begin{subfigure}{.19\linewidth}
        \centering
        \includegraphics[width=\textwidth]{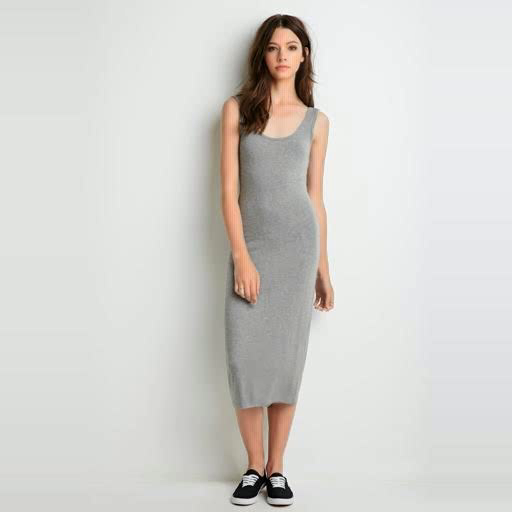}
        \caption{RGB}
    \end{subfigure}
    \begin{subfigure}{.19\linewidth}
        \centering
        \includegraphics[width=\textwidth]{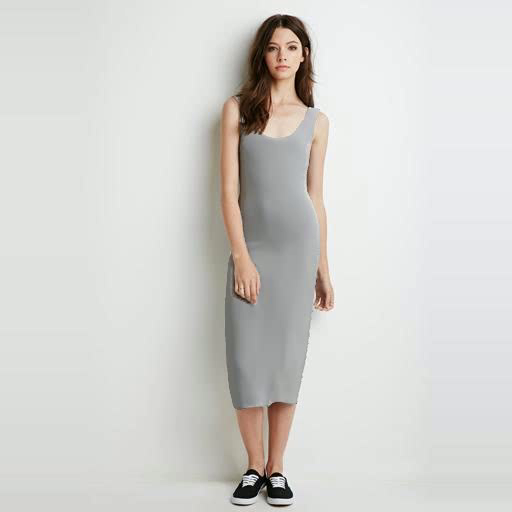}
        \caption{Normals}
    \end{subfigure}
    \begin{subfigure}{.19\linewidth}
        \centering
        \includegraphics[width=\textwidth]{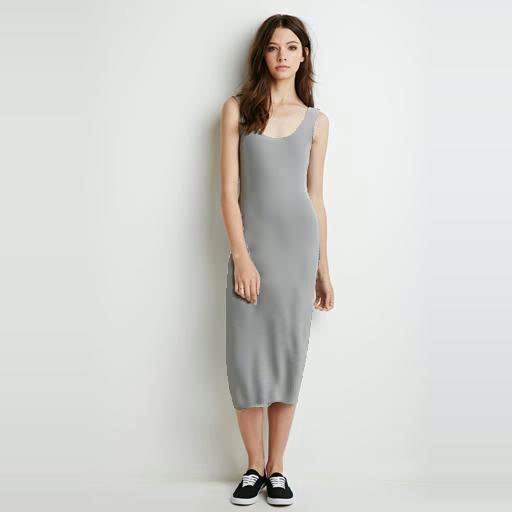}
        \caption{Estimated Normals}
    \end{subfigure}
    \begin{subfigure}{.19\linewidth}
        \centering
        \includegraphics[width=\textwidth]{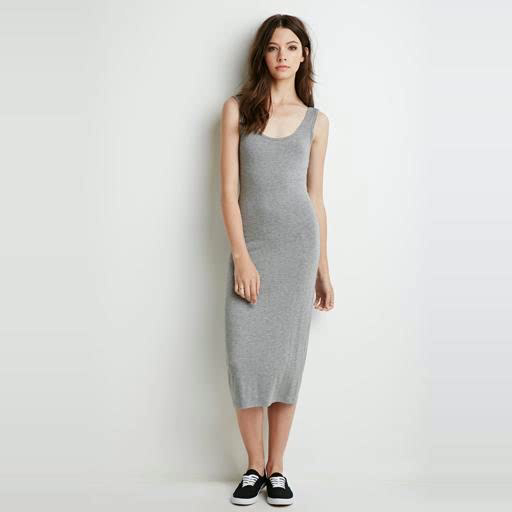}
        \caption{Input}
    \end{subfigure}
    \caption{We analyze the behaviour of each model during generalization using real samples from DeepFashion\cite{liuLQWTcvpr16DeepFashion}. As can be seen, the work of \cite{mahapatra2022controllable} shows color artifacts. Then, our model shows a tradeoff between reconstruction fidelity and wrinkle generation. The RGB solution generates images where color is faithfully maintained, but shows no wrinkles or motion except for very few specific samples. On the other hand, normal-based solutions are not always able to generate the same color distribution due to limitations in the re-shading algorithm (Sec.~\ref{sec:reshading}). Additionally, we observe only the model trained on estimated normals is able to generalize properly to real samples. We omit LIA\cite{wang2022latent} from this comparison as it is unable to generalize at all to real samples.}
    \label{fig:generalization}
    \vspace{-.35cm}
\end{figure*}

\begin{table}[b!]
\renewcommand{\tabcolsep}{0.065cm}
\footnotesize
\centering
\begin{tabular}{lccccccc}
\toprule
 & \multicolumn{5}{c}{Synthetic Data} & \multicolumn{1}{l}{} & \multicolumn{1}{l}{Real Data} \\
Experiment & MAE & MSE & RMSE & SSIM & PSNR & Perceptual & FVD \\
\midrule
LIA\cite{wang2022latent} & 23.77 & 1131.62 & 32.45 & 0.11 & 25.50 & 400.26 & 873.86 \\
Controllable\\ Anim.\cite{mahapatra2022controllable} & 9.37 & 302.26 & 15.89 & 0.48 & 34.49 & \textbf{220.86} & 625.45 \\
\midrule
 & & & & & & & \\
\midrule
Toy & \textbf{6.52} & 197.73 & 12.36 & \textbf{0.61} & \textbf{40.76} & \textbf{218.39} & 643.42 \\
RGB & 7.64 & 175.24 & \textbf{12.21} & \textbf{0.56} & \textbf{37.64} & 230.40 & 623.32 \\
Normals & 15.75 & 563.76 & 22.63 & 0.41 & 32.11 & 231.29 & 633.12 \\
Estimated\\ normals & 17.32 & 674.62 & 24.31 & 0.34 & 31.81 & 231.74 & \textbf{613.26} \\
\bottomrule
\end{tabular}
\caption{Quantitative evaluation of the different methods tested including state-of-the-art baselines, LIA\cite{wang2022latent} and controllable animations\cite{mahapatra2022controllable}, and the different variants of our method trained using different data modalities. Normal-based solutions require a re-shading step (Sec.~\ref{sec:reshading})}
\label{tab:ablation}
\end{table}

\paragraph{Ablation.}
To analyze the effectiveness of the design choices, we test additional baselines which are variants of our method. On one hand, we train our CycleNet using RGB data directly, as opposed to the proposed pipeline in which we operate in the normal space. Next, we analyze the performance of the proposed methodology trained using ground truth normal maps. Nonetheless, there is still a domain gap between synthetic normal maps and normal maps estimated from real data. For this reason, we also train our method using normal maps estimated from synthetic RGB data as input, while using ground truth normal maps as labels. Since our eventual goal is to generate cinemagraphs from real images, we also assess the effectiveness of each approach on real data. Finally, we add as a reference, a \textit{Toy} baseline, in which we \textit{generate} the output cinemagraph $\bm{V}$ for a given input image $\bm{I}$ as $\bm{V} = \{\bm{I}, \bm{I}, ..., \bm{I}\}$. This baseline generates high quality videos since it uses the original image but lacks any motion. This baseline  evaluates how informative the different metrics we use are for our task. 

Tab.~\ref{tab:ablation} shows the quantitative evaluation of the different baselines. Supervised metrics are computed using the test split of our synthetic dataset. Then, to measure quantitatively the capacity to generalize to real data, we test using DeepFashion samples as input and compare to the small set of real samples we have captured using FVD~\cite{unterthiner2018towards}. We show qualitative results of the image synthesis capacity of each approach in Fig.~\ref{fig:ablation} and~\ref{fig:generalization} but refer to the supplementary material to better show the results in motion. 
LIA\cite{wang2022latent} is a solution tailored for human face animation, which exploits some characteristics of the domain, such as the structural and motion similarities of different faces performing the same action/expression (e.g., smiling, talking, etc.). Due to domain differences, we observe its performance is poor when trained on our task. It is unable to produce meaningful motions and often generates an average image for each synthetic animation sequence (best seen in the 2nd and 3rd column in Fig.~\ref{fig:ablation}). This is reflected in the quantitative metrics as well. The work of \cite{mahapatra2022controllable} is designed to work for images of fluids, under the simplifying assumption that a video sequence has a constant flow. While this assumption works well for natural images of fluids, it does not hold for the domain of garments. This method can generate consistent images, but with unrealistic motion and artifacts due to its warping based architecture. We see this in the second sample of Fig.~\ref{fig:ablation}. Furthermore, as expected, when trained with only synthetic data, neither of these solutions are able to generalize to real data. 

\begin{figure}
    \centering
    \begin{subfigure}{\linewidth}
        \centering
        \includegraphics[width=\textwidth]{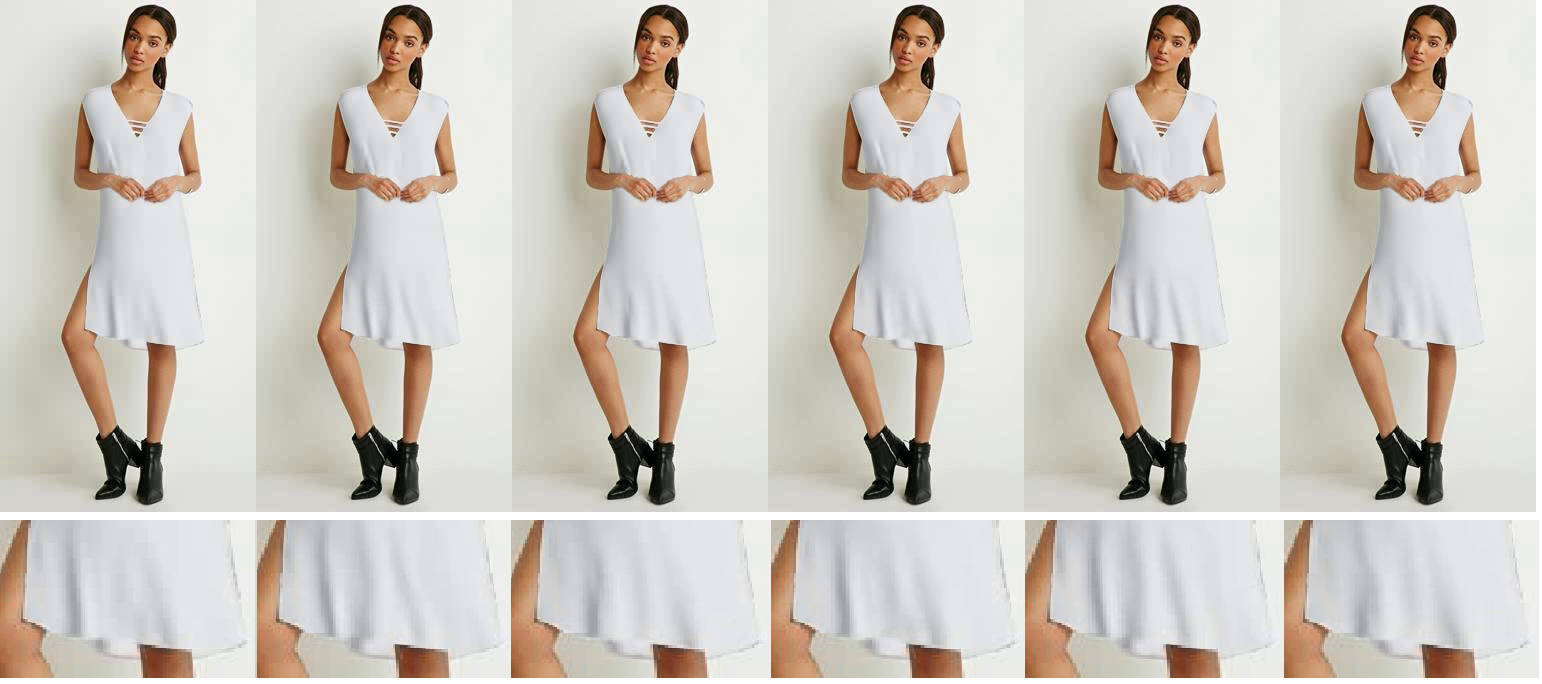}
    \end{subfigure}
    \begin{subfigure}{\linewidth}
        \centering
        \includegraphics[width=\textwidth]{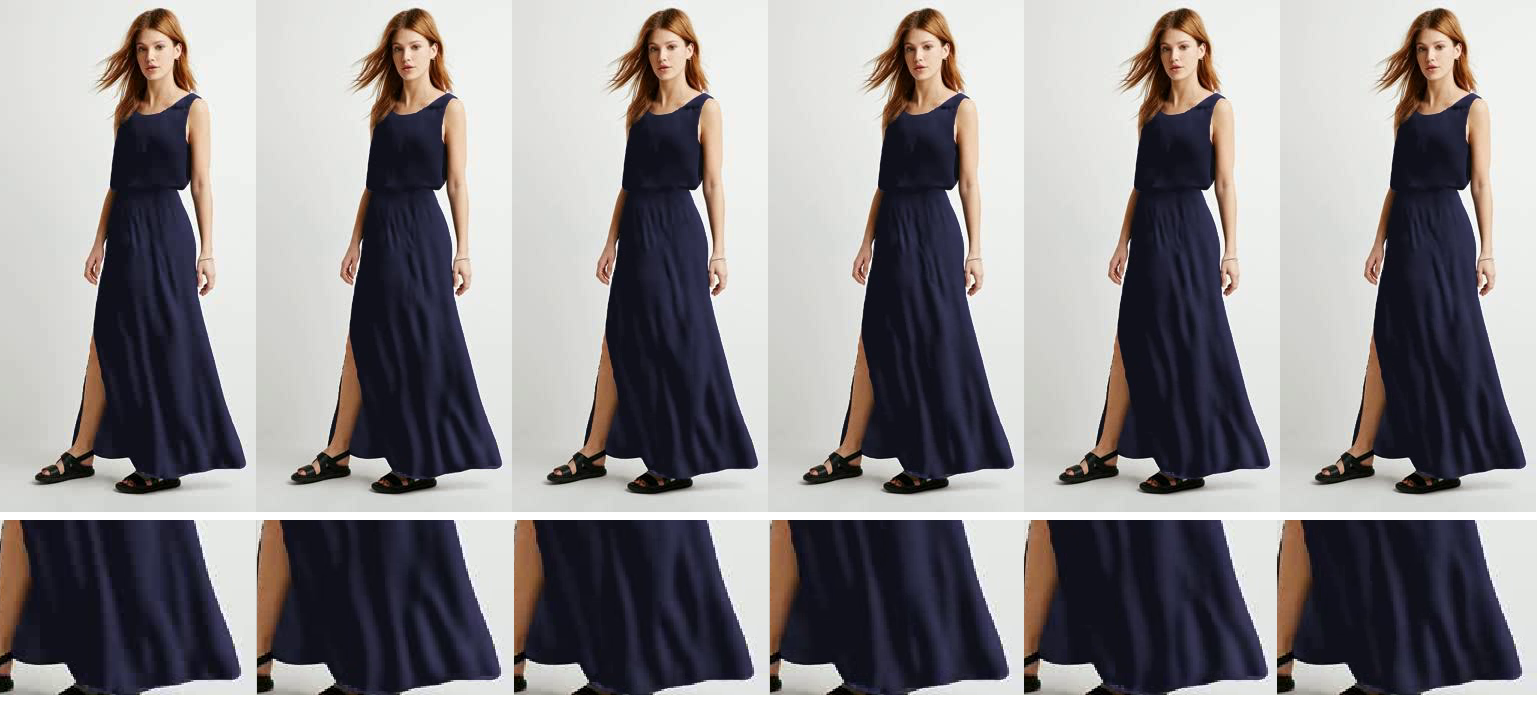}
    \end{subfigure}
    \begin{subfigure}{\linewidth}
        \centering
        \includegraphics[width=\textwidth]{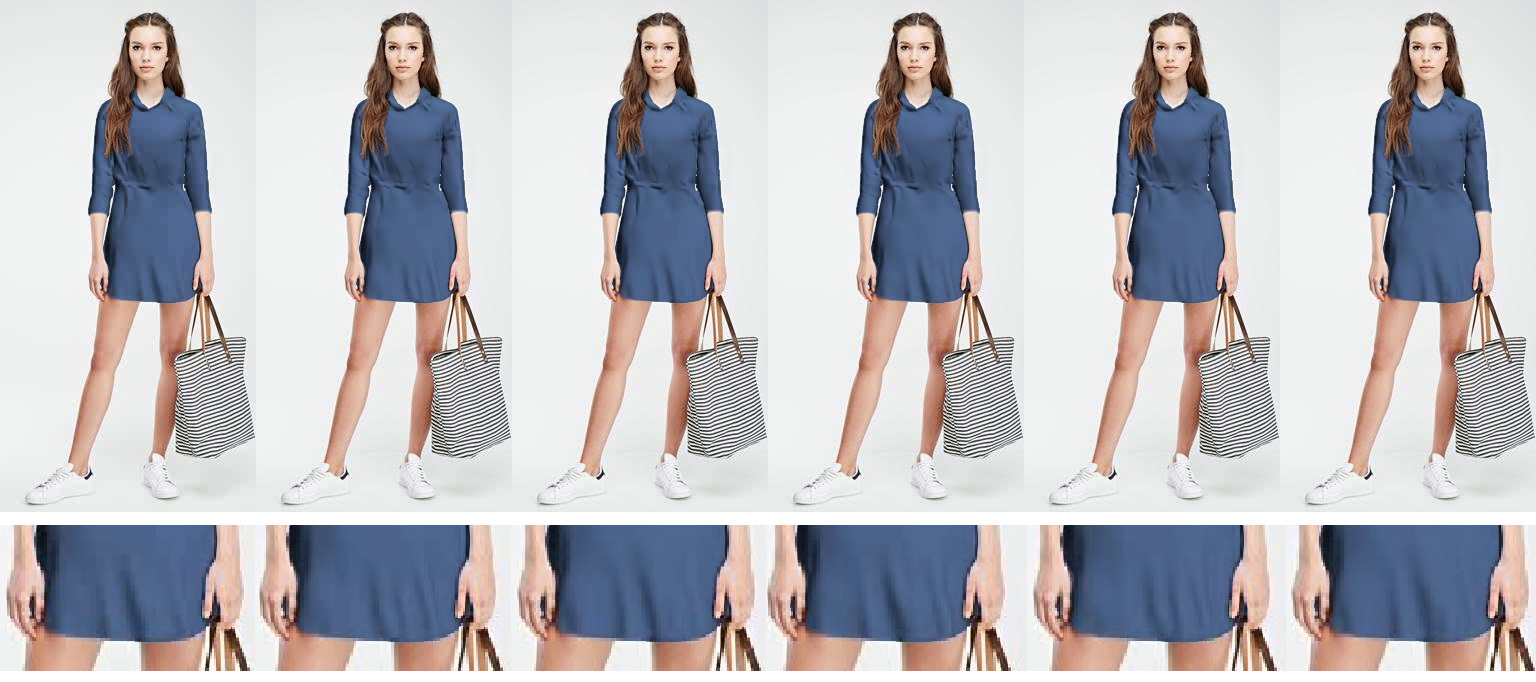}
    \end{subfigure}
    \begin{subfigure}{\linewidth}
        \centering
        \includegraphics[width=\textwidth]{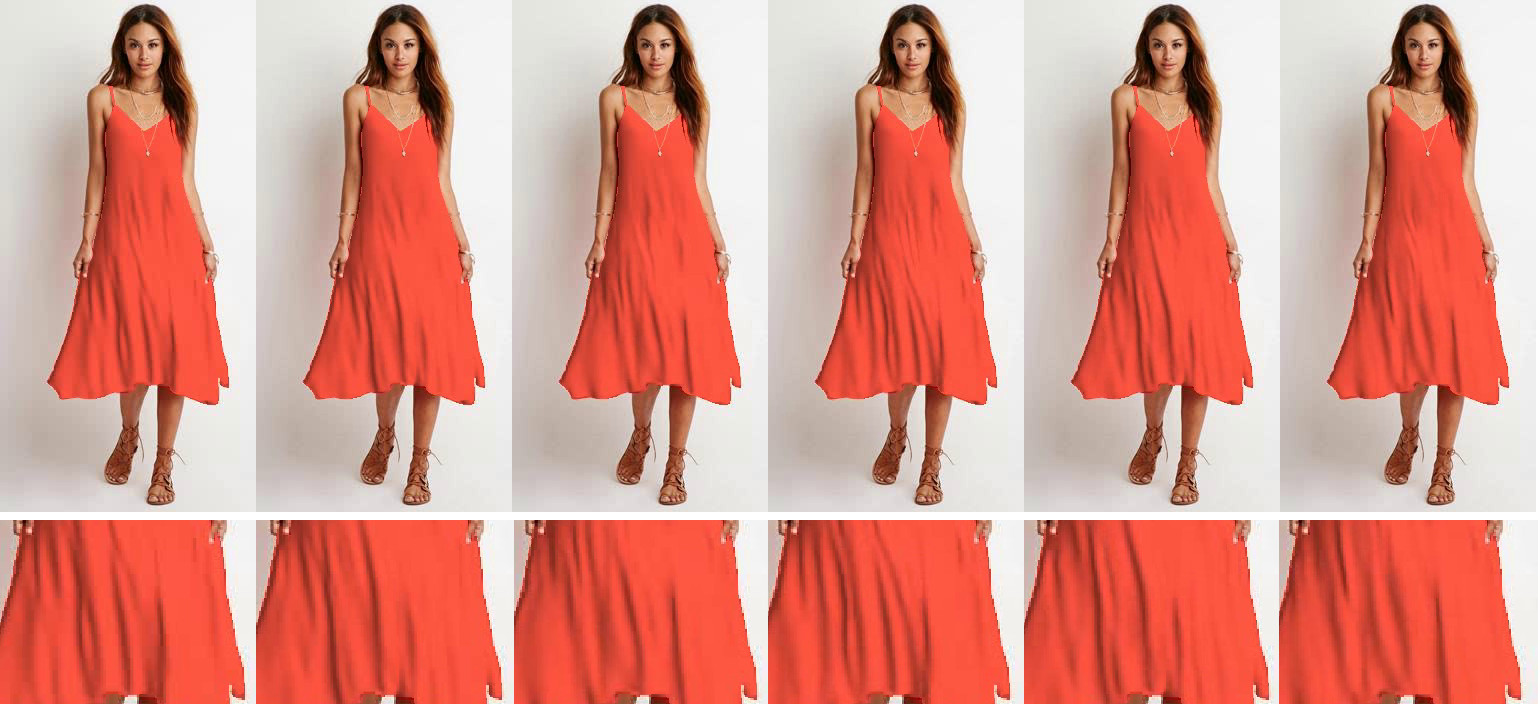}
    \end{subfigure}
    \caption{Qualitative results using real samples from DeepFashion\cite{liuLQWTcvpr16DeepFashion}. We obtain this results with the model corresponding to the last row of Tab.~\ref{tab:ablation}. Our approach is able to generate consistent images with plausible wrinkles. We refer to the supplementary material for a video evaluation of our methodology. We omit LIA\cite{wang2022latent} from this comparison since it is unable to generalize at all.}
    \label{fig:results}
    \vspace{-.35cm}
\end{figure}

\begin{figure}
    \centering
    \includegraphics[width=\linewidth]{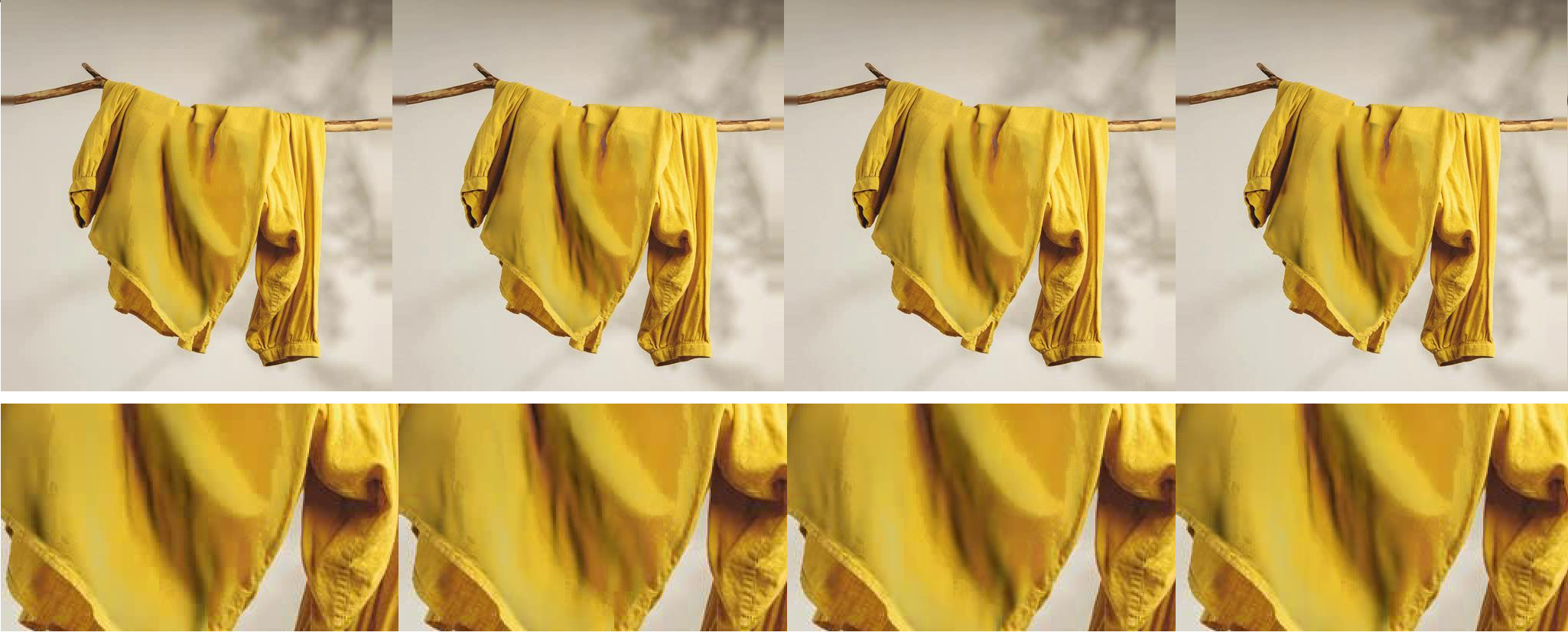}
    \caption{We further test the generalization capacity of our methodology by testing it with an image of a hanging garment. As observed, our approach can synthesize wrinkles on the cloth.}
    \label{fig:test_case}
    \vspace{-.35cm}
\end{figure}

Next, we analyze the different variations of our method. Without a surprise, the RGB solution is the best performing in the synthetic dataset according to the quantitative metrics. Predicting normal map sequences followed by a re-shading step impacts the pixel level reconstruction accuracy as expected. Finally, using estimated normals as input slightly hinders performance w.r.t. using ground truth normals. As observed, the \textit{Toy} baseline performs very well in comparison, while we know it is not generating any motion. This suggests the classical metrics used for image reconstruction have some limitations for evaluating the solutions of our specific problem. The motion we want to generate is localized and subtle. The difference between a plausible motion and a static prediction may be smaller than the reconstruction error of an RGB auto-encoder. 
The solutions based in normal maps require a re-shading step that might produce a slight shift in pixel colors. While this does not hurt the quality of the dynamics, it increases the reconstruction error. The increase in perceptual error from RGB to normal-based solutions is not comparable to that of the other metrics. This suggests that the perceived quality of the generated images is comparable. Next, we evaluate the performance in real samples. For this case, the behaviour we observe is different. While the RGB solution is able to faithfully reproduce the input image with a slight color shift, a large majority of predictions do not show any motion. We observe a similar behaviour with the model trained on ground truth synthetic normals. Due to the domain gap, very few input normal maps produce dynamics in the output. This, added to the color shift due to the re-shading step increases the value of the FVD. Finally, we observe the model trained using predicted normals as input is able to generate plausible dynamics for all the input normals estimated from real samples resulting in the best FVD score. Note how in this case, the \textit{Toy} solution, which \textit{generates} static videos, has the worst FVD score. In, Fig.~\ref{fig:results}, we show additional sequence results for our model trained on estimated normals. For each sequence, we show frames sampled uniformly every $25$ frames. We also add close-up looks of the wrinkles. Our method is able to generate visually appealing results with plausible wrinkles. Finally, we further test the generalization capacity of our method by testing with an image of a hanging garment. This can be seen in Fig.~\ref{fig:test_case}. We refer to the supplementary material for qualitative video results.

Finally, due to the limitations of the quantitative metrics, we complement the evaluation with a qualitative user study. We show random samples of generated animations both on synthetic and real data with the different methods. We ask the users to rate if the animations are plausible or not. Variants of our method are rated as plausible more than $90\%$ of the time on synthetic data. For real images, our method trained on predicted normals is rated as plausible around $60\%$ of the time whereas our method trained on RGB and ground truth normals is rated as plausible around $20\%$ and $30\%$ of the time respectively. The strongest baseline \cite{mahapatra2022controllable} is not perceived as plausible neither on real nor synthetic data. We also ask for an estimation of the perceived wind direction (left or right). Around $\sim70\%$ of users correctly identified the wind direction under which the sequence was generated. We refer to the supplementary material for more details of the user study and additional video evaluation of the wind controllability.

\section{Conclusion}

We introduced a method to generate human cinemagraphs from single RGB images. Our main contribution is a cyclic neural network that produces looping video clips. We demonstrated it is possible to train the network with synthetic data and generalize to real data. To do so, we propose working in the image normal space to close the gap between the different data distributions.

While generating plausible results, our method has some limitations. Intrinsic image decomposition which we use as a step to synthesize back RGB images is a challenging problem and often lacks the high-frequency texture details of the original garments, which are lost during the re-shading step. To address this limitation, in the future we would like to tailor a generic solution towards the type of fabric materials and textures that we are interested. Further, we would like to extend our setup to jointly optimize for normal estimation and motion prediction steps, in an end-to-end fashion. 
Finally, a challenging next step would be to add movement on hair strands that can add significant realism to the cinemagraphs. Since hair simulator is far from being a solved problem, it may be worth rethinking the setup to directly learn a neural (hair) simulator from real video footage, thus closely bringing together research in neural simulators and conditional generative models.

\paragraph{Acknowledgements.} This work has been partially supported by the Spanish projects PID2019-105093GB-I00, TED2021-131317B-I00 and PDC2022-133305-I00 (MINECO/FEDER, UE) and CERCA Programme/Generalitat de Catalunya.) This work is partially supported by ICREA under the ICREA Academia programme. We also thank Aniruddha Mahapatra for his valuable assitance and guidance in the development of this work.

{\small
\bibliographystyle{ieee_fullname}
\bibliography{egbib}
}

\end{document}